\documentclass[runningheads]{llncs}
\usepackage{graphicx}
\usepackage{comment} 
\usepackage{color}
\makeatletter
\newcommand{\figcaption}[1]{\def\@captype{figure}\caption{#1}}
\newcommand{\tblcaption}[1]{\def\@captype{table}\caption{#1}}
\makeatother
\begin{document}
\title{Analyzing Font Style Usage and\\ Contextual Factors in Real Images}%
%\thanks{Supported by organization x.}}
%
% \author{Anonymous submission}
\author{Naoya Yasukochi\inst{1}\and
Hideaki Hayashi\inst{2}\and
Daichi Haraguchi\inst{1}\and
Seiichi Uchida\inst{1}}

\authorrunning{N. Yasukochi et al.}

\institute{Kyushu University, Fukuoka, Japan \and Osaka University, Suita, Japan}
%\email{lncs@springer.com}\and
%ABC Institute, Rupert-Karls-University Heidelberg, Heidelberg, %Germany\\
%\email{\{abc,lncs\}@uni-heidelberg.de}}
%
\maketitle              % typeset the header of the contribution
\begin{abstract}
%The abstract should briefly summarize the contents of the paper in
%15--250 words.
There are various font styles in the world. Different styles give different impressions and readability. This paper analyzes the relationship between font styles and contextual factors that might affect font style selection with large-scale datasets. For example, we will analyze the relationship between font style and its surrounding object (such as ``bus'') by using about 800,000 words in the Open Images dataset. We also use a book cover dataset to analyze the relationship between font styles with book genres. Moreover, the meaning of the word is assumed as another contextual factor. For these numeric analyses, we utilize our own font-style feature extraction model and word2vec. As a result of co-occurrence-based relationship analysis, we found several instances of specific font styles being used for specific contextual factors.
\end{abstract}
%
%
%============================================================
\section{Introduction}
%============================================================
We have various font styles worldwide, giving different impressions and readability. Graphic designers and typographers carefully select appropriate fonts to visualize textual information. For example, when they select a font in a context where some ``stable'' impression is required, they might select a sans-serif font with thick strokes and might not choose a serif font with thin strokes. In this case, some contextual factors determine the fonts selected. \par

% ------------------------------
\begin{figure}[t]
\centering
\includegraphics[width=\textwidth]{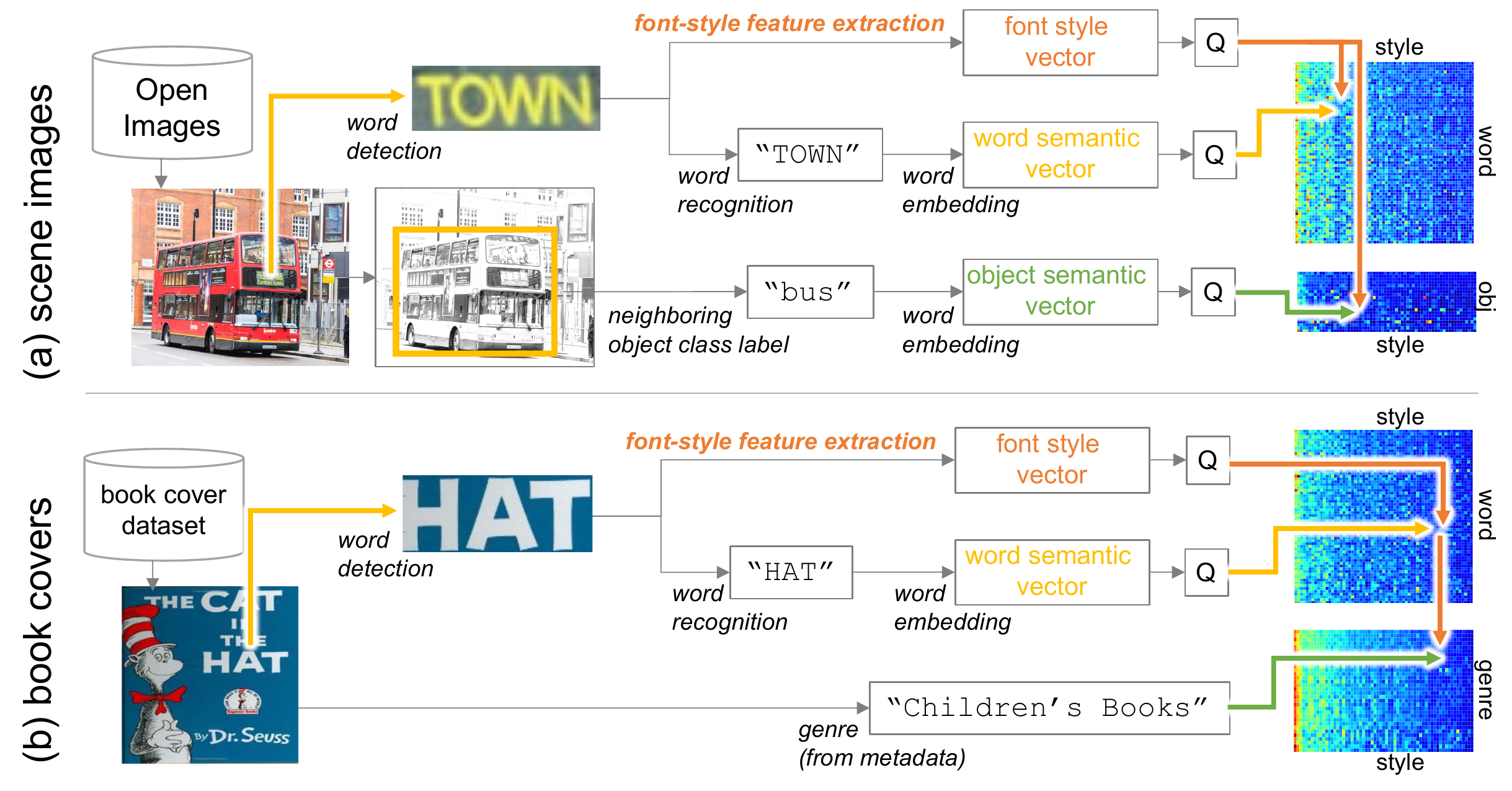}\\[-3mm]
\caption{Two scenarios of font style usage analysis. (a)~Scene images. (b)~Book covers.  `Q' represents quantization by clustering. The rightmost heat maps are the main subjects to be analyzed.}  \label{fig:task}
\end{figure}

The purpose of this paper is to understand the {\em relationship between the style of the selected font and several contextual factors}. Fig.~\ref{fig:task} shows two scenarios for this purpose. The first scenario uses scene images. In (a), the word ``TOWN'' is shown on a bus. In this case, two possible contextual factors that determine the font style of ``TOWN'' are i)~the object ``bus'' where this word is printed and ii)~the meaning of the word ``town.''  The second scenario uses book covers. The style of the word ``HAT'' in (b) might be determined by considering i)~the genre of this book (``Children's Book'') and/or ii)~the meaning of the word ``hat.''  By collecting actual font usage like above and summarizing them as histograms (like heat maps in the rightmost of Fig.~\ref{fig:task}), we will be able to capture the relationship between the font style and those contextual factors. The captured relationship will be useful as empirical knowledge for future typographic designs and, simultaneously, as important statistics to understand human behavior on font selection. \par
However, there are difficulties in the relationship analysis. First, many exceptional cases will hide the weak relationship. For example, the advertisement texts on a bus will not have direct relevance with their contextual factor, the object ``bus.''  Similarly, the word ``Dr.'' in the author's name of the book will have no relationship with the book genre. The font styles of ``Gorgeous hotel'' and ``Cheap hotel'' will be different, although the meaning of the word ``Hotel'' is the same---therefore, in this case, the word meaning is not a strong factor in determining the style. Second, we need to use some numeric ways to handle font styles and word meanings, especially since there is no standard numeric description of font styles.
\par
In this paper, we tackle the first difficulty by using large-scale datasets. For example, in the scene image scenario, we use the Open Images dataset~\cite{OpenImages2}, which contains more than one million images. By using large-scale datasets, we can cancel the perturbations by exceptional cases and will find even a weak trend in the relationship. We also use a simple quantization scheme based on clustering for accumulating weak relationships into stronger ones. For the second  difficulty, we implement a font-style feature extractor based on a convolutional neural network (CNN) and prepare a numeric style vector for each word. Similarly, we use word2vec to have a semantic representation of each word.\par
The main contributions of this paper are summarized as follows.
\begin{itemize}
    \item To the best of the authors' knowledge, this is the first attempt to analyze the word-wise relationship between font style and contextual factors (object around the word (e.g., ``bus''), genre, and meaning of the word) with large-scale datasets.
    \item We set two very different scenarios with scene images and book cover images because they provide different contextual factors. 
    \item The co-occurrence-based analysis with scene images revealed that font styles in scene images showed co-occurrences with objects surrounding the word. For example, handwriting styles and whiteboards tend to co-occur. 
    \item Our analysis of the book cover dataset revealed numerous co-occurrences between font styles and words/genres, with consistent tendencies between word-style and genre-style relationships. For example, the ``Mystery, Thriller, \& Suspense'' genre tends to use condensed sans-serif fonts that co-occur with suspenseful words.
\end{itemize}
\par

%============================================================
\section{Related Work}
\subsection{Font Style Embedding and Application}
Regarding font-related tasks, many font style embedding and extraction approaches have been proposed. Here, we survey various font-related tasks and their font style embedding and extraction approaches.\par

Visual font recognition (VFR) is a task that aims at automatic recognition of the fonts in real images. Chen~\textit{et al.}~\cite{chen2014large} are the pioneer in this task and proposed a handcrafted font embedding approach based on the local image descriptors using max pooling. Following this, DeepFonts~\cite{wang2015deepfont}, a deep learning-based approach using a convolutional neural network (CNN) to recognize fonts, was proposed. Wang~\textit{et al.} also tackled the VFR task for Chinese fonts using CNNs.\par

Font retrieval and recommendation are also hot topics in font-related tasks~\cite{o2014exploratory,choi2019assist,kulahcioglu2020fonts,matsumura2020font}. To retrieve or recommend fonts, query font images or tags associated with fonts are mainly used. Several researchers employed deep CNNs such as ResNet-18 and ResNet-50 to embed font styles into a feature space~\cite{choi2019assist,kulahcioglu2020fonts,matsumura2020font}. In particular, \cite{kulahcioglu2020fonts,matsumura2020font} used the font style and word embeddings by word2vec~\cite{Mikolov2013EfficientEO,mikolov2013distributed} and achieved font retrieval and recommendation from query font images and tags.\par

In addition, we introduce three font-related tasks in real images. For web designs, a task aiming at estimating a font that matches a target web design has been proposed~\cite{zhao2018modeling}. As a task using movie posters, a novel task that estimates time periods of the release date of a movie by using fonts on a query movie poster has been tackled~\cite{tsuji2021using}. Jiang~\textit{et al.}~\cite{jiang2019visual} proposed a method for automatically identifying header and body font pairs from PDF pages. In all the above three tasks, CNNs were used to extract and embed font styles.\par

As we reviewed above, deep learning-based approaches were mainly used to embed font styles regardless of the tasks. Following the existing trials, we also used a deep learning model to extract font style features.

\subsection{Multimodal Analysis of Fonts}
Kulahcioglu~\textit{et al.} conducted two interesting multimodal analyses related to fonts~\cite{kulahcioglu2018predicting,kulahcioglu2019paralinguistic}. One is the analysis of the relationship between fonts and semantic attributes, \textit{e.g.,} ``angular,'' ``artistic,'' and so on~\cite{kulahcioglu2018predicting}. The other is the analysis of the effectiveness of fonts in word clouds~\cite{kulahcioglu2019paralinguistic}.\par

Shinahara~\textit{et al.}~\cite{shinahara2019serif} analyzed the correlation between fonts in book covers, their colors, and the genres of books. Their analysis is similar to ours; however, there are clear differences. First, they classified font images into predefined six font styles by simply matching query font images to reference font images, whereas we obtained font style embeddings by using a CNN, and then we conducted clustering of the style embeddings to 64 clusters. From this difference, we can analyze the relationship between fonts and contextual factors by paying attention to more detailed styles. Second, we do not address font colors yet analyze the correlation between fonts and word semantics in book covers. Our study aims to understand the trend of font style usage; therefore, we use semantics that might correlate with font styles.

\subsection{Word Embedding and its Multimodal Analysis}
Word embedding, a technique that transforms the semantics of a word into a numerical vector, constitutes one of the fundamental methodologies in recent natural language processing. Among the variety of word embedding methods, word2vec~\cite{Mikolov2013EfficientEO,mikolov2013distributed} stands out as the most prominent one. The underlying principles of word2vec include two core techniques: continuous bag-of-words and Skip-gram. Both methods are based on the distributional hypothesis that words appearing in the same sentence are semantically related. In the learning process of Skip-gram, the surrounding words of the target word are inferred from the target word, while the target word is assumed to be inferred from the input of the surrounding words in the learning process of continuous bag-of-words.\par

In addition, a pre-trained natural language processing model known as bidirectional encoder representations from transformers (BERT)~\cite{devlin2018bert} has exhibited superior performance over word2vec and thus has been increasingly used in recent studies. The advantage of BERT is its capability to address ambiguity. While word2vec provides a unique word embedding for a given input, BERT generates diverse word embeddings for the same word depending on the context.\par

Several types of research use word embedding for multimodal analysis between words and other modalities~\cite{ikoma2020effect,takeshita2021label}. 
In this paper, we employ word2vec to obtain static word embeddings for analyzing the co-occurrence between words and font styles. This allows us to quantitatively analyze the relationship between words and font styles.

%============================================================

%============================================================
\section{Analysis of Font Style Usage in Real Images}
%============================================================
%------------------------------------------------------------
\subsection{Experimental Strategy} 
%------------------------------------------------------------
This study aims to reveal how font styles are used in real-world images. In particular, it is considered useful to investigate the co-occurrence of font styles with written words and contextual factors in the images, such as neighboring objects in scene images and metadata in document images. A straightforward way to analyze this co-occurrence involves detecting the text in the image and identifying the font using a classifier. However, a perfect font classifier is difficult to attain, while relatively accurate font classifiers such as DeepFont~\cite{wang2015deepfont} have been proposed. In addition, a font classifier needs to be fine-tuned to the dataset, but image datasets that contain both font labels and rich meta-data are unavailable.\par

Therefore, we employed a semi-manual strategy based on feature extraction and clustering. After detecting texts in the real images and cropping them as image patches, we first extract features from the image patches using a CNN so that fonts can be well separated in the feature space. We then perform clustering in the feature space. We manually analyze the font family of each cluster and examine the co-occurrence of the font styles and contextual factors for each cluster. This process inevitably produces clusters containing random fonts, which are manually discarded.\par 

The experiment consists of the following three parts: (a) verification of the trained style feature extractor, (b) analysis of font style usage in scene images, and (c) analysis of font style usage in book cover images. In part (a), we trained a style feature extractor with a synthetic font image dataset and evaluated it using another font image dataset with font labels. Part (b) involves analyzing how font styles are used in scene images according to words and neighboring objects. Part (c) investigates font usage in book cover images based on words and book genre. 
% Details on datasets, the style feature extractor, scene text detection and recognition, and word embedding are described in the subsequent subsection.

%------------------------------------------------------------
\subsection{Image Dataset}
\subsubsection{AdobeVFR Real~\cite{wang2015deepfont}}
AdobeVFR real is a large-scale real-world text image dataset. This dataset contains 4,384 text images collected from various typography forums, which covers 617 font classes. Each image has a hand-annotated font label inspected by independent experts. 
% Images are cropped with tight bounding boxes and normalized proportionally in size to be with an identical height of 105 pixels. 
AdobeVFR real is provided as a subset of the AdobeVFR dataset, which includes an unlabeled font image dataset and synthetic font image dataset in addition to AdobeVFR real. Considering the purpose of our experiment, we only employed AdobeVFR real, which is a real-world and labeled image dataset.\par

In the verification of the style feature extractor using AdobeVFR real, we preprocessed the images according to the following procedure: First, we resized each image to be with a height of 64 pixels. If the width was smaller than 128 pixels, the same images were concatenated horizontally until the width reached 128 pixels or greater. When inputting the images into the style feature extractor, we cropped images with a size of 64 pixels $\times$ 128 pixels from the concatenated images at a random location.

\subsubsection{Open Images v4~\cite{OpenImages2}} 
Open Images v4 (hereinafter referred to as Open Images) is a large-scale scene image dataset annotated with image-level labels, object bounding boxes, and visual relationships. This dataset contains 14,610,229 object bounding boxes for 600 object classes. We filtered out images without bounding boxes from the original dataset, which consisted of approximately 9M images, resulting in a final set of 1,743,042 images.\par

We used this dataset to analyze how fonts are used in the scene images according to the words and the objects neighboring the words. In this analysis, we detected texts in the image and investigated the relationship between used font styles and words/objects. 
% The details for the text detection are described in the following subsection.

\subsubsection{Book Cover Dataset~\cite{shinahara2019serif}} 
This dataset contains 207,572 book cover images collected from Amazon.com along with their corresponding metadata, such as titles and genres. This dataset covers 32 genres, and each image belongs to a single genre. 

We used texts, including book titles, to analyze the use of font styles and their relationship to words and book genres. After detecting and cropping the word on book covers as a patch image, images with a height of smaller than 15 pixels were discarded. We compared the text recognition results with the title information in the meta-data of the book cover and selected words that match the titles. We then excluded stop words, numbers, and compound words. We standardized the spelling of words, such as tenses, using a lemmatizer. 
To eliminate special proper nouns and coined words, we employ words included in the British national corpus (BNC)\footnote{http://www.kilgarriff.co.uk/bnc-readme.html}. Eventually, the number of words used in this experiment resulted in 4,092.

%------------------------------------------------------------
\subsection{Text Detection and Recognition}
\label{sec:text_detection_method}
For text detection and recognition for the Open Images and book cover datasets, we followed the procedures adopted in~\cite{takeshita2021label}. We employed CRAFT~\cite{baek2019character} to detect texts in the images. CRAFT provides word bounding boxes, which are rotatable, thereby allowing the detection of rotated texts. For character recognition, we used a combination of models called ``TPS-ResNet-BiLSTM-Attn,'' which is proposed in~\cite{baek2019character}. This model employs thin-plate spline (TPS) removing spatial distortion, ResNet, BiLSTM, and attention-based sequence prediction (Atten) for feature extraction, sequence modeling, and word recognition.

%------------------------------------------------------------
\subsection{Font Style Feature Extraction}
\label{sec:feature_extractor}
The CNN model for extracting font style features (hereinafter, referred to as a style feature extractor) is constructed by training a CNN as a font classifier. A CNN trained as a classifier is supposed to acquire an embedding from an input image space into a feature space in which features are separable for each class. This nature can be used to obtain a feature embedding suitable for clustering. This idea of using a font classifier as a feature extractor for clustering is inspired by the method in~\cite{jiang2019visual}.\par

We employed ResNet-18~\cite{he2016deep} as the backbone model for the style feature extractor. We modified the network structure by inserting a fully-connected layer with a 200-dimensional output and a ReLU activation function between the average pooling layer and the final layer of the original model to obtain 200-dimensional style features whose dimension is the same as that of the text feature. We also applied dropout~\cite{srivastava2014dropout} to the added fully-connected layer to improve the generalization capability. After the network is trained as a classifier, we removed the final layer and treated the output from the added fully-connected layer as the style feature.\par

\begin{figure}[t]
 \begin{center}
\includegraphics[keepaspectratio,width=\textwidth]{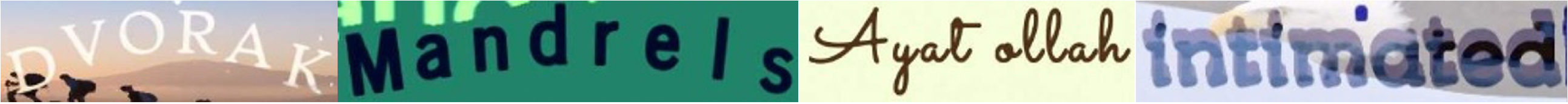}\\[-3mm]
\caption{Examples from the image dataset for training the style feature extractor. Images are generated by SynthTIGER~\cite{yim2021synthtiger}.}  \label{fig:synthtiger_examples}
\end{center} 
\end{figure}

We synthesized a text image dataset using SynthTIGER~\cite{yim2021synthtiger} and used it as a training dataset for the style feature extractor. Image examples from our dataset are shown in Fig.~\ref{fig:synthtiger_examples}. We selected 200 fonts in Google Fonts\footnote{https://fonts.google.com} and generated 1,000 images for each font; namely, 200,000 images were obtained in total. Words were randomly selected from the corpus of MJsynth. To ensure the diversity of fonts, 40, 40, 40, and 80 fonts were randomly selected from serif, sans-serif, handwriting, and display fonts, respectively. We generated images with a height of 64 pixels while maintaining the aspect ratio; if the width was smaller than 128 pixels, the same images were concatenated horizontally until the width reached 128 pixels or greater\footnote{We tried padding to unify the width of the images, but this concatenation-based method was better for style feature extraction}. When inputting the images into the style feature extractor, we cropped images with a size of 64 pixels $\times$ 128 pixels from the original images at a random location to unify image sizes and improve the generalization capability. To avoid non-italic fonts becoming italicized, we removed the process of tilting texts from the original implementation.\par 
% ==============================================
% ==============================================

In the training of the style feature extractor, the adam optimizer~\cite{kingma2014adam} with a learning rate of 0.001 was employed. The batch size was set to 64, and the maximum number of training epochs was set to 100 with early stopping. 

%------------------------------------------------------------

%------------------------------------------------------------
\subsection{Word Embedding}
\label{sec:word2vec}
We employed word2vec~\cite{mikolov2013distributed} to vectorize the object class names for the Open Images dataset and detected words for both datasets. Word2vec is a well-known word embedding method that can represent each word as a 200-dimensional semantic vector used in \cite{takeshita2021label}. 

%============================================================
\section{Experimental Results}
%============================================================
%------------------------------------------------------------
\subsection{Verification of the Trained Style Feature Extractor} 
\label{sec:VFR_experiment}
The purpose of this experiment is to demonstrate the validity of the style feature extractor explained in Section~\ref{sec:feature_extractor} using a dataset with font names as ground truth. A style feature extractor trained with SynthTIGER is expected to produce features that separate styles well. However, it is unknown whether the extractor will perform similarly on datasets other than SynthTIGER.\par 

Therefore, we evaluate the feature extractor using the following procedure: First, we extract style features from the AdobeVFR real dataset using the style feature extractor trained with SynthTIGER. We then perform clustering on the obtained style features. The clustering result is qualitatively and quantitatively evaluated by visualizing the extracted features via $t$-SNE and calculating the adjusted mutual information (AMI) between the clusters and the font labels. The AMI is a variation of mutual information that measures the strength of the relationship between two indices while correcting the effect of the random agreement. If this index is sufficiently higher than zero, the clustering result is at least not random with respect to the ground truth; thus, feature vectors that reflect font style information are supposed to be extracted. The numbers of clusters were set to $64$ and $614$, which are the number of clusters to be used in the following experiments and the ground-truth number of classes, respectively.

\begin{figure}[t]
    \centering
    \includegraphics[keepaspectratio,scale=0.32]{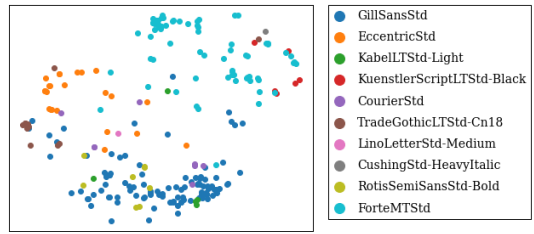}\\[-3mm]
    \caption{Visualization of the style features of ten randomly selected fonts from the Adobe VFR dataset based on $t$-SNE.} \label{fig:tsne_VFR}
\end{figure}

Fig.~\ref{fig:tsne_VFR} visualizes the style features of the AdobeVFR real dataset extracted by the style feature extractor. To enhance clarity, ten fonts were randomly selected from the original dataset of 614 fonts, as the complete visualization became excessively cluttered. This figure shows that the style features are separated for each font to some extent, thereby suggesting that the features are clusterable.
% ==============================================
% ==============================================

The AMI scores are $0.2645$ and $0.1590$ when the numbers of clusters are 64 and 614, respectively. The scores are sufficiently higher than zero, and thus there is a non-random relationship between the clustering results and ground truth of font labels, suggesting that the extracted features reflect the font style information.

%------------------------------------------------------------
\subsection{Analysis of Font Style Usage in Scene Images} 
%------------------------------------------------------------
\label{sec:SceneImageAnalysis}
To reveal how font styles are used in real images according to words and surrounding objects, we analyzed the Open images dataset. As the foundational information of this dataset, the histogram of the top 30 frequent words is shown in Table~\ref{table:OpenImageWord_top30}. It should be noted that words that are not included in BNC are excluded because the original list of the detected words includes many meaningless such as a single letter of the alphabet. Table~\ref{table:OpenImageObject_top30} shows the top 30 frequent objects in the Open Image dataset. The total number of detected object types was 578.\par

%======word frequent table==============
\begin{table}[t]
\caption{Top 30 frequent words in Open Images Dataset. Words are displayed only if they are included in the British national corpus (BNC).}\vspace{-3mm}
\label{table:OpenImageWord_top30}
\resizebox{\textwidth}{!}{%
\begin{tabular}{lllll}
\hline
1-10th: `new,' `photography,' `one,' `state,' `park,' `make,' `go,' `world,' `get,' `day' \\
10-20th: `stop,' `open,' `city,' `free,' `time,' `police,' `life,' `may,' `service,' `good' \\
20-30th: `first,' `house,' `work,' `school,' `love,' `photo,' `big,' `way,' `book,' `use' \\
\hline
\end{tabular}%
}
\end{table}

%======object frequent table==============
\begin{table}[t]
\caption{Top 30 frequent objects in Open Images Dataset.}\vspace{-3mm}
\label{table:OpenImageObject_top30}
\resizebox{\textwidth}{!}{%
\begin{tabular}{lllll}
\hline
1-10th: `poster,' `building,' `clothing,' `book,' `man,' `person,' `tree,' `car,' `woman,' `bottle' \\
10-20th: `house,' `bus,' `plant,' `truck,' `drink,' `girl,' `boat,' `van,' `beer,' `train' \\
20-30th: `table,' `boy,' `wine,' `camera,' `television,' `window,' `suit,' `box,' `food,' `toy' \\
\hline
\end{tabular}%
}
\end{table}

% **Procedure**
The procedure of this experiment is as follows. First, we detect scene texts in individual images as described in Section~\ref{sec:text_detection_method} and cropped them as patches. We then extract style features using the style feature extractor, which is explained in Section~\ref{sec:feature_extractor}. We also extracted word features based on word2vec as explained in Section~\ref{sec:word2vec}. The analysis based on the extracted features is two-fold: word--style analysis and object--style analysis. The details of each analysis are described below. We used $k$-means clustering, and the number of clusters in each clustering trial was determined based on the gap statictics~\cite{tibshirani2001estimating}.

\subsubsection{Word--Style Analysis}
\label{sec:openimages_word-sytle}
In this analysis, we examine the relationship between words and font styles detected in the images. First, we performed clustering using the word features to examine what kind of word groups are created. Second, we also clustered the style features to confirm what style features are contained in the dataset. Finally, we analyzed the co-occurrence between words and styles by visualizing the heat map of two feature vectors.\par 

\begin{figure}[t]
 \centering
  \includegraphics[keepaspectratio, scale=0.5]{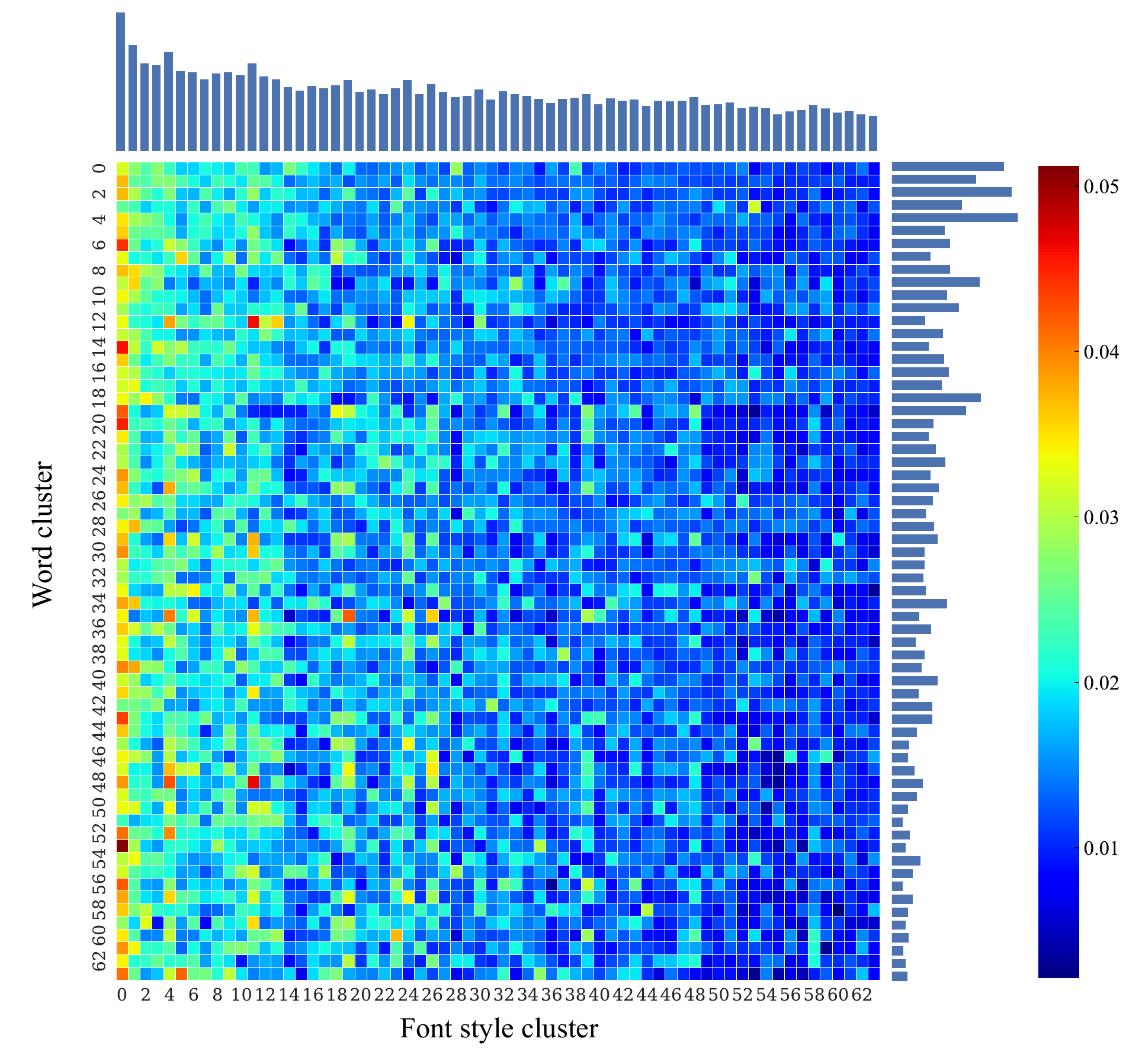}
  \\[-5mm]
\caption{Heat map of word clusters and font style clusters for Open Images.}\label{fig:OpenImg_StyTxt}
\end{figure}

% **Cooccurrence between text and style **
Fig.~\ref{fig:OpenImg_StyTxt} shows a heat map of font style clusters and word clusters for Open Images. To avoid meaningless co-occurrences caused by multiple occurrences of images with the same word in the same font, such as repeated logos, we considered any word that was significantly larger in scale (specifically, more than 10 times the median) than the number of occurrences of other words, as a single instance in each cluster. The heat map was normalized for each row by dividing the value in each cell by the total row sum. The procedures of avoiding meaningless co-occurrences and normalization are consistently applied to all subsequent heat maps.
% ==============================================
% ==============================================

Table~\ref{table:word-style_pairs_OpenImg} shows the significant combinations identified in Fig.~\ref{fig:OpenImg_StyTxt}. There were not so many co-occurrences between words and font styles, but we found some characteristic co-occurrences that suggest the relationship between words and font style usage. 
% ==============================================
\begin{table}[t!]
\caption{Characteristic co-occurrences and their details in Fig.~\ref{fig:OpenImg_StyTxt}.}\vspace{-3mm}
\label{table:word-style_pairs_OpenImg}
\resizebox{\textwidth}{!}{%
\begin{tabular}{lllll}
\hline
Word cluster ID & Style cluster ID & Members & Estimated meaning & Font styles \\
\hline
12 & 4, 11, 13 & `much,' `ordinary,' `little,' `many,' `big' & Adjective \& Adverb& Serif \\
19 & 5, 18, 19 & `service,' `software,' `digital,' `retail,' `technology' & IT \& Service & Sans-serif \\
23 & 21, 22, 23 & `station,' `seat,' `vehicle,' `bus,' `tunnel' & Vehicles & Sans-serif \\

\hline
\end{tabular}%
}
\end{table}
% ==============================================
Our findings in this analysis are summarized as follows:
\begin{itemize}
    \item Adjectives and adverbs tend to be written in serif fonts.
    \item IT and service-related terms tend to be written in sans-serif fonts.
    \item Vehicle-related terms tend to be written in sans-serif fonts.
\end{itemize}

% *Cooccurrence between object and style **
\subsubsection{Object--Style Analysis}
We investigated how font styles are used in scene images according to the objects surrounding the words, we analyzed the co-occurrence of objects and font styles. We conducted this analysis using two approaches: object-wise and object cluster-wise. In the object-wise approach, we examined the co-occurrence of font style clusters with individual names of the top 40 most frequently occurring objects. In the object cluster-wise approach, we analyzed the co-occurrence of font style clusters with clusters of objects that share similar meanings. To obtain object clusters, we performed clustering on the object features, which are the word2vec embedding of object names, and obtained 64 clusters.

However, due to a significant imbalance in the number of members and total instances for each cluster, we excluded clusters with the number of instances below the median and used the remaining 20 clusters for the analysis. 

\begin{figure}[t]
 \centering
\includegraphics[keepaspectratio, scale=0.5]{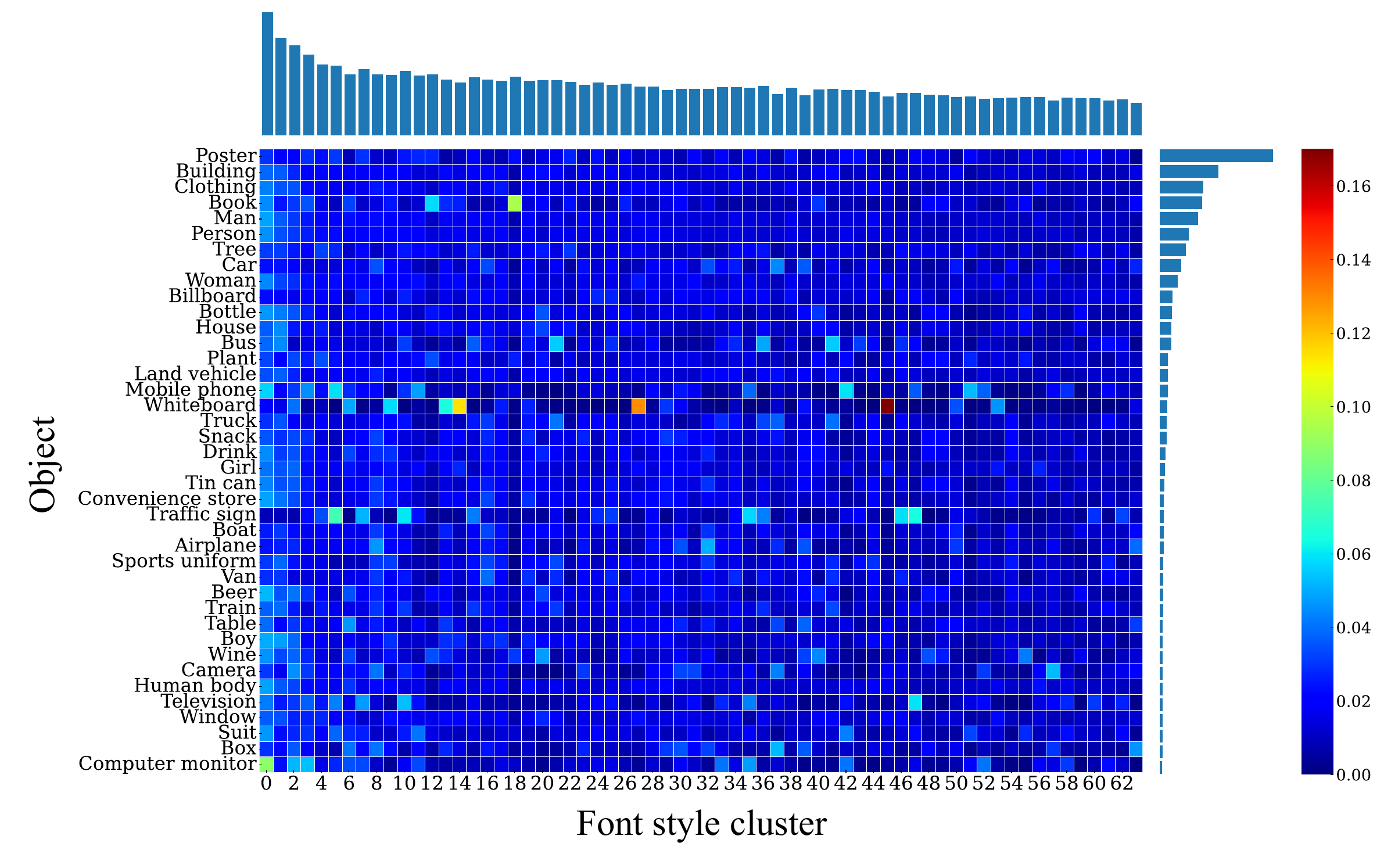}\\[-5mm]
\caption{Heat map of font style clusters and object names for the Open Images dataset.}  \label{fig:OpenImg_StyObjlabel}
\end{figure}

Fig.~\ref{fig:OpenImg_StyObjlabel} shows a heat map of the object names and font style clusters.
There were more co-occurrences than in the word--style relationship. Some characteristic co-occurrences are summarized in Table~\ref{table:object-style_pairs_OpenImg}.
% ==============================================
% ==============================================
% ==============================================
\begin{table}[t]
\caption{Characteristic combinations and their members in Fig.~\ref{fig:OpenImg_StyObjlabel}.}\vspace{-3mm}
\label{table:object-style_pairs_OpenImg}
\centering
\begin{tabular}{lllll}
\hline
Object label & Style cluster ID & Font styles \\
\hline
Book & 12, 18 & Serif \\
Bus & 21, 36, 41 & Sans-serif condensed \\
& & Sans-serif with narrow kerning \\
Whiteboard & 14, 27, 45 & Handwriting \\
% Mobile phone & 7, 42, 51 & Sans-serif \\
Mobile phone & 7, 42, 51 & Sans-serif\\
Traffic sign & 7, 10, 47 & Sans-serif \\
Television & 10, 47 & Sans-serif\\
Computer monitor & 33, 35, 42 & Sans-serif \\
\hline
\end{tabular}%
\end{table}
% ==============================================

Based on these results, our findings are summarized as follows:
\begin{itemize}
    \item Texts on books tend to be written in serif fonts.
    \item Texts on buses involve sans-serif type fonts such as sans-serif condensed and sans-serif with narrow kerning.
    \item Characteristic fonts that co-occur with mobile phones are sans-serif, sans-serif regular, and sans-serif thin; text written on mobile phones tends to be in sans-serif type fonts.
    \item Texts written on whiteboards tend to be in cursive style. 
    \item Traffic signs tend to co-occur with sans-serif fonts.
    \item Words nearby televisions are written in sans-serif fonts.
    \item Computer monitors often involve sans-serif fonts.
\end{itemize}

\begin{figure}[t]
 \centering
  \includegraphics[keepaspectratio, scale=0.5]{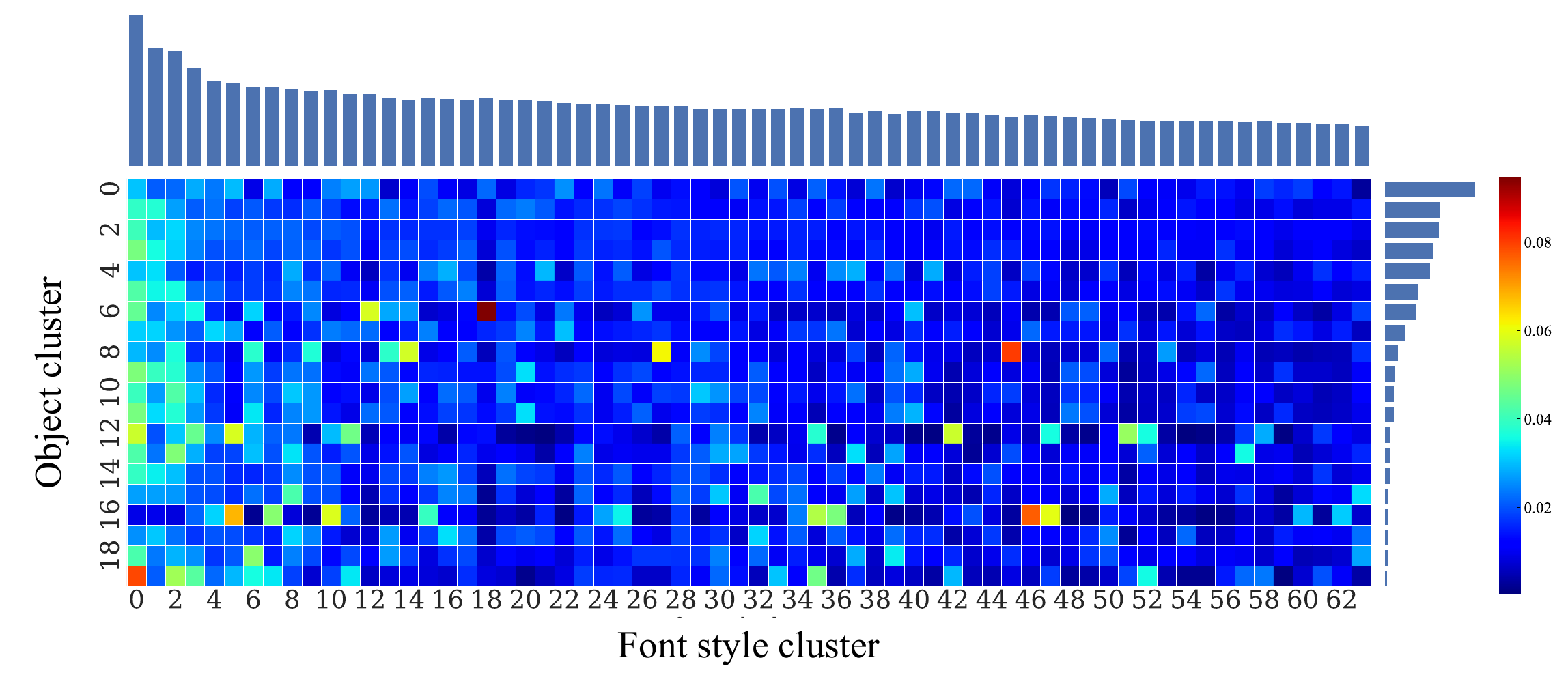}\\[-5mm]
\caption{Heat map of font style clusters and object clusters for the Open Images dataset.}  \label{fig:OpenImg_StyObj}
\end{figure}

Fig.~\ref{fig:OpenImg_StyObj} shows a heat map of the object clusters and font style clusters, and Table~\ref{table:objectcluster-style_pairs_OpenImg} summarizes noteworthy co-occurrences. 
% ==============================================
% ==============================================
While there were only a few explicit instances of co-occurrence, some remarkable patterns emerged. For example, clusters containing words related to house appliances tended to use sans-serif fonts. 
% ==============================================
\begin{table}[t]
\caption{Characteristic co-occurrences and their details in Fig.~\ref{fig:OpenImg_StyObj}.}\vspace{-3mm}
\label{table:objectcluster-style_pairs_OpenImg}
\resizebox{\textwidth}{!}{%
\begin{tabular}{lllll}
\hline
Object cluster ID & Style cluster ID & Members & Estimated meaning & Font styles \\
\hline
12 & 5, 42, 51 & `Mobile phone,' `Corded phone,' `Telephone' & Phone & Sans-serif \\
16 & 10, 46, 47 & `Traffic sign,' `Stop sign,' `Tick' & Sign & Sans-serif \\
19 & 11, 35, 52 & `Computer monitor,' `Home appliance,' `Calculator' & House appliances & Sans-serif \\
\hline
\end{tabular}%
}
\end{table}
% ==============================================

%------------------------------------------------------------
\subsection{Analysis of Font Style Usage in Book Cover Images} 
%------------------------------------------------------------

% Purpose
We investigated how font styles are used in book cover images according to words and genres. 
% **Procedure**
We employed a similar procedure to that described in Section~\ref{sec:SceneImageAnalysis}. We detected texts in the title region in each book cover image and cropped them as patches. We then extract style features using the style feature extractor and word features based on word2vec. Besides word--style analysis the same as in Section~\ref{sec:SceneImageAnalysis}, we also conducted genre--style analysis using the pre-defined genres of the book cover dataset. 

\subsubsection{Word--Style Analysis}
In this analysis, we examined the relationship between words and font styles detected in the images. We performed clustering on the word features and style features and then analyzed the co-occurrence between words and styles by visualizing the heat map of two feature vectors. We performed clustering on the word features and style features and obtained 62 and 63 clusters respectively. 

\begin{figure}[t]
 \centering
\includegraphics[keepaspectratio, scale=0.51]{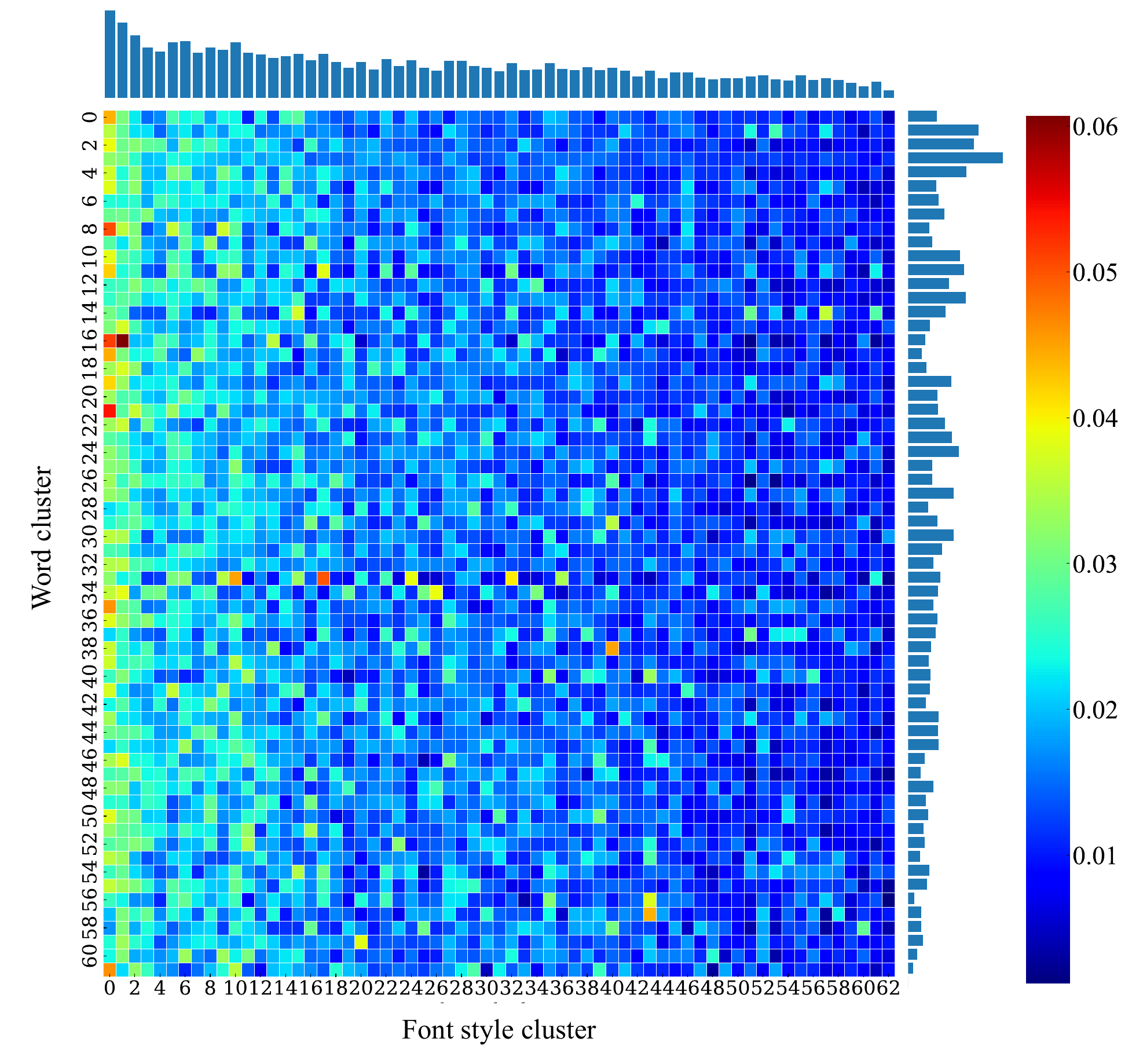}\\[-5mm]
\caption{Heat map of the word clusters and font style clusters in the book cover dataset}  \label{fig:Book_cover_StyTxt}
\end{figure}

% **Cooccurrence between text and style **
Fig.~\ref{fig:Book_cover_StyTxt} shows a heat map of font style clusters and word clusters for the book cover dataset. 
% ==============================================
% ==============================================
% ==============================================
\begin{table}[t]
\caption{Characteristic co-occurrences and their details in Fig.~\ref{fig:Book_cover_StyTxt}.}\vspace{-3mm}
\label{table:word-style_pairs_book}
\resizebox{\textwidth}{!}{%
\begin{tabular}{lllll}
\hline
Word cluster ID & Style cluster ID & Members & Estimated meaning & Font styles \\
\hline
8 & 5, 9 & `book,' `poetry,' `novel,' `diary' & Press cover & Serif \\
11 & 9, 10, 17 & `empire,' `king,' `queen,' `castle' & Aristocracy & Serif \\
33 & 10, 17, 32 & `prayer,' `sacred,' `god,' `saint' & Religion & Serif \\
34 & 25, 26, 34 &  `computer,' `application, `software,' `data' & IT & Sans-serif \\
40 & 27, 35, 43 & `kill,' `crime,' `murder,' `spy' & Suspense & Sans-serif condensed\\
46 & 11, 27, 39 & `battle,' `war,' `violence,' `army' & War & Sans-serif condensed\\
58 & 10, 16, 25 & `gene,' `molecular,' `serum,' `chemical' & Chemistry & Serif, Sans-serif \\
\hline
\end{tabular}%
}
\end{table}
% ==============================================
Table~\ref{table:word-style_pairs_book} summarizes characteristic co-occurrences in Fig.~\ref{fig:Book_cover_StyTxt}. Our findings in this analysis are summarized as follows:
\begin{itemize}
    \item Religion-related words tend to be written in serif fonts.
    \item IT-related words tend to be written in sans-serif fonts.
    \item Aristocracy-related words tend to be written in serif fonts.
    \item Although words related to the press cover a wide range of fonts, they are relatively often written in serif.
    \item Words used in suspense stories such as kill, suicide, and murder are all written in sans-serif condensed.
    \item War-related words are often written in sans-serif condensed, as well as suspense stories.
    \item Words related to chemistry and medicine tend to be written in serif and sans-serif fonts.
\end{itemize}

% \item Vehicle-related words are often written in sans-serif semi-bold.
% txt46は乗り物系の単語が多いクラスタであり、特徴的な点は（sty8,31,39)であり、sans serif semi-boldがふくまれるフォントクラスタであり、乗り物系の単語には少し太めのサンセリフ体が用いられる傾向がある。

\subsubsection{Genre--Style Analysis}
We investigated how font styles are used in book cover images according to the book genre. We examined the co-occurrence of the book genres and font styles by visualizing the heat map of them. Thirty-two genres provided in the book cover dataset are used. We used the clustering results of the style features in the word--style analysis above.  

\begin{figure}[t]
 \centering
  \includegraphics[keepaspectratio, scale=0.4]{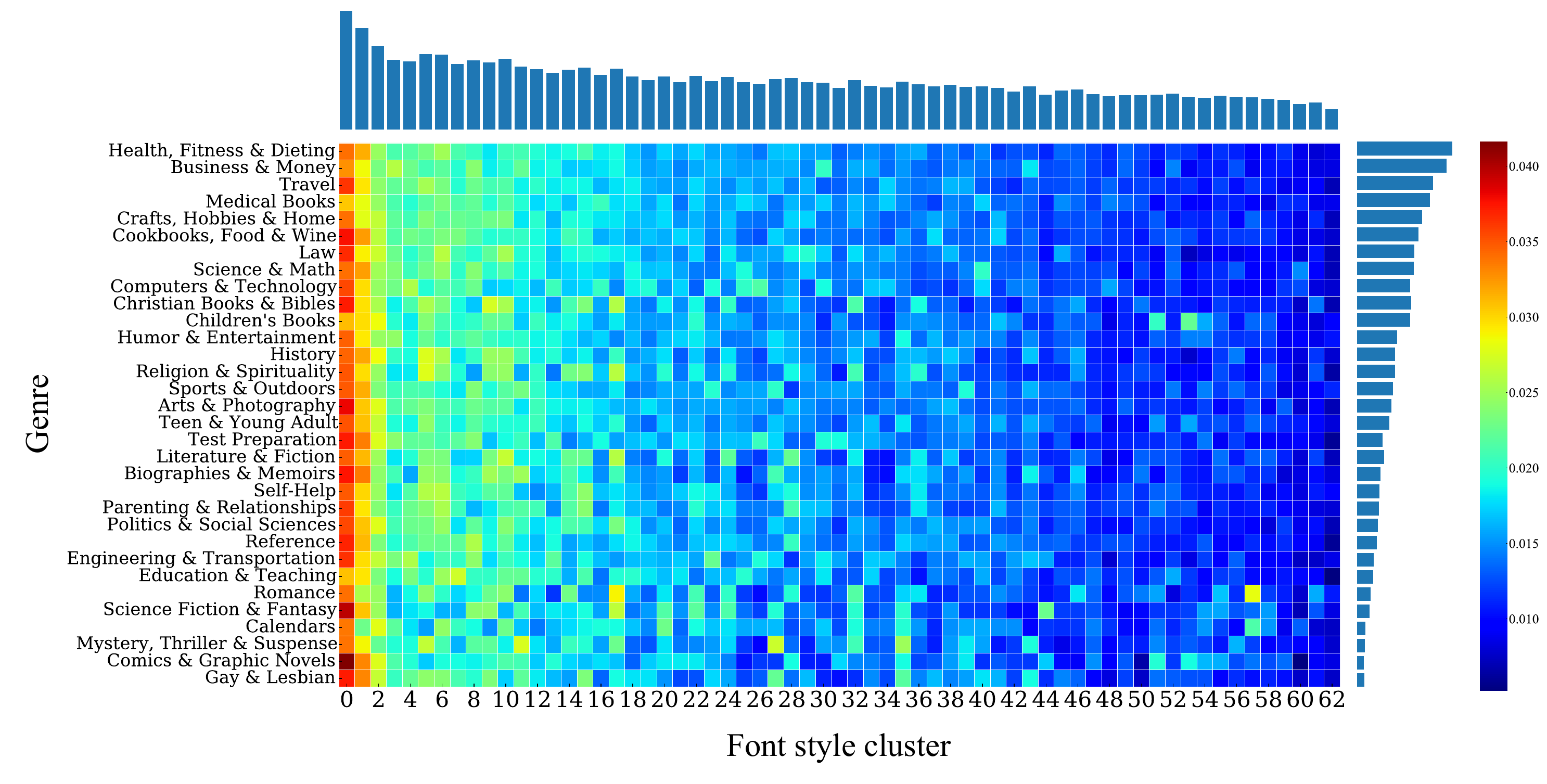}\\[-5mm]
\caption{Heat map of the genres and font style clusters in the book cover dataset}  \label{fig:Book_cover_StyGenre}
\end{figure}

Fig.~\ref{fig:Book_cover_StyGenre} shows a heat map of genres and style clusters for the book cover dataset. 
% In this figure as well, meaningless co-occurrences that have arisen due to multiple occurrences of the same images are excluded beforehand. 
% ==============================================
% ==============================================
% ==============================================
\begin{table}[t]
\caption{Characteristic co-occurrences and their details in Fig.~\ref{fig:Book_cover_StyGenre}.}\vspace{-3mm}
\label{table:genre-style_pairs_OpenImg}
\centering
\begin{tabular}{lllll}
\hline
Genre & Style cluster ID & Font styles \\
\hline
Law & 6, 10 & Serif \\
Computers \& Technology & 4, 8, 13 & Sans-serif \\
Christian Books \& Bibles & 9, 15, 17 & Serif \\
Children's Books & 51, 53 & Handwriting,  Fancy \\
History & 5, 6, 9 & Serif \\
Religion \& Spirituality & 5, 10, 17 & Serif \\
Sports \& Outdoors & 11, 23, 27 & Sans-serif \\
Literature \& Fiction & 10, 17, 24 & Serif \\
Engineering \& Transportation & 4, 8, 23 & Sans-serif \\
Mystery, Thriller \& Suspense & 11, 27, 35 & Sans-serif condensed \\
\hline
\end{tabular}%
\end{table}
% ==============================================
Our findings in this analysis are summarized as follows:
\begin{itemize}
    \item ``Law,'' ``Christian Books \& Bibles,'' ``History,'' ``Religion \& Spirituality,'' and ``Literature \& Fiction'' include many serif fonts.
    \item ``Business \& Money,'' ``Computers \& Technology,'' ``Sports \& Outdoors,'' ``Test Preparation,'' ``Engineering \& Transportation,'' ``Education \& Teaching,'' ``Mystery, Thriller \& Suspense'' often involve sans-serif fonts.
    \item ``Children's Books'' and ``Comics \& Graphic Novels'' use handwriting and fancy fonts more than other genres.
\end{itemize}
It is noteworthy that some findings in Table~\ref{table:genre-style_pairs_OpenImg} are consistent with those observed in the word-style analysis in Table~\ref{table:word-style_pairs_book}. Serif fonts are frequently used in ``Christian books \& Bibles,'' which suggests a strong correlation with the finding that religious word clusters co-occur with serif fonts. Additionally, the ``Mystery, Thriller, \& Suspense'' genre tends to use condensed sans-serif fonts that co-occur with suspenseful words.

%============================================================
\section{Conclusion}
%============================================================
In this paper, we aimed at understanding the relationship between the style of the selected font and several contextual factors. For this purpose, we conducted a co-occurrence-based analysis of font styles and contextual factors such as used words, surrounding objects, and book genres for scene images and book cover images.\par 

In the analysis of font styles in scene texts using the Open Images, there were not so many co-occurrences between words and font styles; however, we found some interesting co-occurrences such as adjectives and adverbs which are often written in serif fonts. For objects and font styles, we found many co-occurrences. For example, text on traffic signs tends to be written in sans-serifs, texts on book pages tend to be written in serifs, and mobile phones, television, and computer monitor tend to be written in sans-serifs\par

As for the book cover, we found many co-occurrences between words and font styles more than in the Open Images. The clusters of religious words tended to use serif fonts, while the clusters of IT words tended to use sans serif fonts. In addition, the clusters of suspense and war-related words tended to use a condensed sans-serif font. In terms of genres and font styles, we found that fonts such as handwriting and fancy tend to be used more in children's books than in other genres. We also found some similarities between word--style analysis and genre--style analysis. For example, the ``Christian books \& Bibles'' genre often uses serif fonts that co-occur with the religious word cluster, and the ``Mystery, Thriller \& Suspense'' genre frequently involves condensed sans-serif fonts that co-occur the suspenseful word cluster.\par

In future work, we will conduct a composite analysis using multiple contextual factors. For example, we can calculate the conditional probabilities $p(\mathrm{style} \mid \mathrm{contextual~factor})$ and investigate which factor strongly affects the usage of font styles. We will also conduct experiments on other types of datasets such as document images. 

%============================================================
\section*{Acknowledgment}
%============================================================
This work was supported in part by JSPS KAKENHI Grant Numbers JP22H00540 and JP21H03511.
%
% ---- Bibliography ----
%
% BibTeX users should specify bibliography style 'splncs04'.
% References will then be sorted and formatted in the correct style.
%
\bibliographystyle{splncs04}
\bibliography{main}
%

% \begin{thebibliography}{8}
% \end{thebibliography}

\end{document}